\pdfoutput=1
\documentclass{article}

\usepackage{PRIMEarxiv}

\usepackage[utf8]{inputenc}
\usepackage[T1]{fontenc}

\usepackage{booktabs}
\usepackage{amsfonts}
\usepackage{nicefrac}
\usepackage{microtype}
\usepackage{lipsum}
\usepackage{fancyhdr}
\usepackage{graphicx}
\usepackage{enumitem}
\graphicspath{{media/}}
\usepackage[backend=biber,style=numeric,sorting=none]{biblatex}
\usepackage{xurl}
\usepackage[breaklinks=true]{hyperref}

\title{Multinational AGI Consortium (MAGIC):\\ A Proposal for International Coordination on AI}

\author{
  Jason Hausenloy\thanks{Equal contribution of all authors. Andrea Miotti is the corresponding author.} \\
  UWCSEA East \\
  \texttt{hause56773@uwcsea.edu.sg} \\
  \And
  Andrea Miotti\footnotemark[1] \\
  Conjecture \\
  \texttt{andrea@conjecture.dev} \\
  \And
  Claire Dennis\footnotemark[1] \\
  Princeton University \\
  \texttt{claire.dennis@princeton.edu} \\
}

\begin{document}
\maketitle

\begin{abstract}
This paper proposes a Multinational Artificial General Intelligence Consortium (MAGIC) to mitigate existential risks from advanced artificial intelligence (AI). MAGIC would be the only institution in the world permitted to develop advanced AI, enforced through a global moratorium by its signatory members on all other advanced AI development. MAGIC would be exclusive, safety-focused, highly secure, and collectively supported by member states, with benefits distributed equitably among signatories. MAGIC would allow narrow AI models to flourish while significantly reducing the possibility of misaligned, rogue, breakout, or runaway outcomes of general-purpose systems. We do not address the political feasibility of implementing a moratorium or address the specific legislative strategies and rules needed to enforce a ban on high-capacity AGI training runs. Instead, we propose one positive vision of the future, where MAGIC, as a global governance regime, can lay the groundwork for long-term, safe regulation of advanced AI. 
\end{abstract}

\section{Executive Summary}

Today, a handful of U.S.-based AI companies are spearheading the development of extremely powerful, general AI systems – what the leading developers refer to as Artificial General Intelligence (AGI). AGI lacks any unified, concrete definition, but has been described in some contexts as “superintelligence” or “highly autonomous systems that outperform humans at most economically valuable work.”\cite{openai_charter} While the specific threshold for defining an “AGI” is often contested, there is an emerging scientific and public consensus that the unchecked development of such advanced, autonomous AI systems may pose enormous, societal-scale risks.\cite{chan_harms_2023,ngo_alignment_2023,carlsmith_2022} The challenges posed by advanced AI are sizable enough to necessitate an international response. A growing chorus of policymakers, technologists, and governance experts are thus calling for global AI governance through various proposed international bodies to facilitate global coordination and oversight.\footnote{ These include but are not limited to: UN Secretary-General Antonio Guterres, Microsoft CEO Satya Nadella, OpenAI CEO Sam Altman, UK Prime Minister Rishi Sunak (in conversation with US President Joe Biden) and Chair of the UK Foundation Model Taskforce Ian Hogarth.}\cite{un_guterres,microsoft_ceo,openai_superintelligence,uk_pm,uk_taskforce}  

Some recent AI governance proposals rely on \textit{post-hoc} auditing and verification mechanisms – evaluating the system once it has already been developed. \cite{model_evaluation} Given the potential of advanced AI systems to rapidly improve during training, this method is insufficient. Regulation of most global hazards, such as biological weapons, do not require the technology to be first engineered in order to subsequently control it. To intervene in the pre-development stage, however, requires centralized, tightly controlled, and constant oversight of advanced AI models, with methods in place to abort training if there are signs of dangerous capabilities. To comprehensively halt unchecked AGI advancement therefore requires monopolization of development in a centralized facility, coupled with a simultaneous moratorium on all unofficial development. 

To achieve this level of oversight, this paper proposes an \textbf{exclusive Multinational AGI Consortium (MAGIC)} – an international governance structure designed to avoid existential risks of AI systems. MAGIC has the following four core characteristics:
\begin{enumerate}
    \item \textbf{Exclusive}: the world’s only advanced AI facility, with a monopoly on the development of advanced AI models, and nonproliferation of AI models everywhere else.
    \item \textbf{Safety-focused}: focused on the development of AI systems that are safe by design, including development of new architectures and ways to bound existing AI systems.
    \item \textbf{Secure}: among the most highly secure facilities on Earth, with strict protocols for information security.
    \item \textbf{Collective}: supported internationally, where the benefits of AI systems are distributed among all member countries.
\end{enumerate}

\section{Multinational AGI Consortium (MAGIC)}

The Multinational AGI Consortium (MAGIC ) relies on four core characteristics: 1) exclusivity, 2) focus on safety, 3) security, and 4) collective action. MAGIC goes further than other proposals for international AI research institutions in its call for an immediate restriction of all external advanced AI development.

\subsection{Exclusive: The World’s Only Advanced AI Facility}

\textbf{MAGIC would be the sole institution authorized to conduct advanced AI research beyond a defined capability level.} In this single, secure facility, the world's leading AI experts would research advanced AI systems and new architectures to design powerful, safe AI systems. MAGIC would be the only organization authorized to train advanced AI models above a certain threshold of computing power. It would attract top global talent through its exclusive resource access, cutting-edge research, reputation, and funding.\footnote{ Other large-scale scientific collaborations similarly maintain control over specific technologies through the physical concentration of resources. CERN hosts the world's largest particle physics laboratory, including the Large Hadron Collider, and retains exclusive access to these facilities. \cite{CERN_2022_2023} ITER similarly hosts the world's largest experimental fusion reactor in France which cannot be replicated by any single nation due to its cost and complexity. \cite {ITER} \cite{ITER_cyber}}

\textbf{This exclusivity would make any advanced AI development outside of MAGIC illegal, enforced through a global moratorium on training runs using more than a set amount of computing power.}\footnote{We suggest that one way to implement such thresholds would be to establish two thresholds: a lower “Frontier Threshold” and a higher “Horizon Threshold”. These would be calculated in terms of total computation used for an AI model’s training run (measured in \texttt{FLOP}) and would help classify AI training runs and apply adequate measures to each:
\begin{enumerate}
\setlength{\itemsep}{-0.05ex}    
  \setlength{\parskip}{-0.01ex}    
    \setlength{\parsep}{-0.01ex}     

    \item Training runs that meet or exceed the ``Frontier Threshold'' must adhere to mandatory oversight requirements from the national jurisdiction where they are being conducted in, including closer monitoring, stringent reporting, and regular audits.
    \item Training runs that meet or exceed the ``Horizon Threshold'' should be prohibited by default.
\end{enumerate}
These are discussed in a proposal for the priorities of the UK Foundation Models Taskforce, and Dalrymple elaborates on how such thresholds could be defined and calculated. \cite{Miotti_2023,davidad}}\footnote{In some versions, MAGIC could provide additional computational power remotely through a 'compute bank'. This would be a centralized resource of computing power that can be accessed remotely. The operations would primarily monitor resource usage and applications of this model. Finally, AI safety research, external to MAGIC, would focus on fundamental technical problems.  These include the mechanistic interpretability of models, formal proofs of boundedness, and ultimately, value alignment.}\cite{anderljung_frontier_2023} Signatory countries to MAGIC would mandate cloud-computing providers, with whom almost all AI companies partner, to prevent any training runs above a specific size within their national jurisdictions. \footnote{We suggest $10^24$ floating point operations (FLOP) as an appropriate threshold given current levels of compute usage in frontier models \cite{owid_computation}} Signatory countries would be responsible for ensuring mutual verification of the enforcement of this measure. \footnote{This would be similar to some “IAEA-for-AI” models, which serve as a standard and compliance enforcement role through inspecting hardware, which may require direct physical access to data centres, or model evaluations after development to test the competencies of the model. \cite{Hausenloy_Dennis_2023, baker_nuclear_2023}} Domestically, this would not require the creation of any additional monitoring mechanisms, as cloud providers already track the size of training runs to ensure compliance with their terms of service and for routine business operations.\footnote{Data centers are a key entry point for regulators. They could be required to disclose large training runs which could then be audited by an international regulatory regime.\cite{shavit_2023} \cite{Pilz}}

\textbf{To make the moratorium sustainable in the long-term, MAGIC members would have to further monitor and control hardware to train AI systems.} AI accelerators, advanced chips such as Nvidia’s H100s specialized for AI applications, are currently available for purchase to most commercial actors worldwide and are untracked.\footnote{ This is true apart from a handful of entities under export controls by the US government.}\cite{shavit_2023} There are very few actors who have the expertise and financial resources to aggregate sufficient compute to train state-of-the-art models.\footnote{Compute researcher Lennart Heim estimates there are “less than 100 actors, probably…30 actors” who could aggregate enough compute to achieve this.\cite{heim_2023}} However, accessibility will increase as enormous amounts of capital continue to flow into AI development and available chips become more powerful and less expensive. This moratorium could be supplemented with stricter monitoring of GPU supply chains.\footnote{ In the long run, implementing this would be harder. To maintain a moratorium, a coalition of actors would have to intervene in global semiconductor supply chains to restrict their ability to obtain or build the capacity to manufacture AI accelerators. Members achieve this by imposing export controls on the AI accelerators, their design blueprints, and the equipment needed for their production.}\cite{baker_nuclear_2023} MAGIC member states would intervene in global semiconductor supply chains to restrict the ability to obtain or build the capacity to manufacture AI accelerators, for example by building a ‘chip directory.’\cite{shavit_2023} Members achieve this by imposing export controls on AI accelerators, their design blueprints, and equipment needed for their production for non-signatory states or non-compliant signatory countries

\begin{center}
    \fbox{
        \parbox{0.8\textwidth}{
            \textbf{Why monitor and restrict compute?}

            Compute is an effective lever for centralized intervention. Compared to alternatives such as algorithms or data, compute is a physical resource that is easily detectable and already monitored. As the large training runs required for frontier model development utilize vast quantities of these chips, compute measurement provides a proxy to measure the capabilities of models for regulators. \cite{openai_compute, sevilla_2022, henshall_2023} Most importantly, while the data and algorithms used for the most powerful AI models might vary over time, all AI and all computation will always need computing power, and the amount of computing power will likely continue to be correlated to the power of the AI system. It is universal and agnostic to AI architecture. A cap on compute affects most compute-intensive modern architectures, such as reinforcement learning and deep learning and non-transformers such as diffusion models. This approach is therefore effective in the short-term until significant improvements in algorithmic efficiency necessitate adjustments. 
        }
    }
\end{center}

\textbf{\textit{Limitations:}} Enforcing exclusivity, and a moratorium, through compute requires that computing power remains a strong proxy for how advanced an AI model is. In general, scale factors like compute and dataset size are still expected to remain highly relevant to pushing the capabilities of frontier models, but incentives are driving algorithmic efficiency advances which could begin to overturn this assumption.\footnote{OpenAI in 2020: “Our results suggest that for AI tasks with high levels of investment (researcher time and/or compute) algorithmic efficiency might outpace gains from hardware efficiency (Moore’s Law).” \cite{openai_efficiency}}\cite{openai_efficiency, CSET_compute_2023} Additionally, some research indicates that, in any case, the current rate of scaling may not be viable in the long-term. If this is the general trend, compute governance may not continue to be the most direct proxy for model size in the long term.\cite{CSET_compute_2023} In other words, as advanced models become smaller and cheaper to build, establishing a compute-based threshold for monitoring the most powerful AI systems may become unviable. MAGIC’s response to account for this should be to lower the threshold over time, while investing in safety research such that the highest available capabilities threshold is being used, coupled with a monitoring agency to ensure that the best safety techniques are being followed. Flexibility must be built into MAGIC’s monitoring metrics to account for algorithmic progress over time and ensure it is nimble enough to incorporate new capabilities.
\subsection{Safety-Focused: Differential Technological Development of Safe Architectures}

\textbf{MAGIC's core mandate will not be to build the most powerful AI models as soon as possible, but to build them as safely as possible.} MAGIC would  encourage the differential technological development of architectures that are safer while restricting the training of AI through today's black-box, foundation model method.

\textbf{MAGIC would move from a paradigm in which safety is an afterthought, to safety being built into the design from the beginning.} This would begin by experimenting with the largest current models (for example, GPT-4 level). Instead of automatically scaling to the next model, MAGIC would pursue an approach that tweaks and improves safer architectures, making them more interpretable and bounded. MAGIC would first focus on developing models that are on par with existing capabilities and provably safe.  If MAGIC is successful, it will then move on to slowly expanding its models to be more capable. If it is not proven safe, MAGIC would stop building and wait for new evidence and breakthroughs before restarting its development efforts. 

\textbf{This is important because we do not understand the internal computations of current “black-box” foundation model architectures.\footnote{Prof. Stuart Russell said in a recent Senate Judiciary hearing, of the current paradigm of developing advanced AI models: “They are black boxes…their internals are largely impossible to understand.” \cite{russell_testimony}}\cite{russell_testimony}}They often demonstrate unexpected capabilities, some of which only become evident after deployment to millions of users.\footnote{  Anthropic’s CEO said that,  “You have to deploy [the model] to a million people before you discover some of the things that it can do.”
} Advanced AI models can thus be considered to have “unbounded” capabilities – we cannot be certain what the limits of any system may be, or how to align these systems to human values and goals. MAGIC circumvents these dilemmas by designing advanced AI to be safe from the ground up, rather than having safety be treated as a post-hoc patch.

\textbf{\textit{Limitations:}} These safer approaches, however, come with a tradeoff, or a ``safety tax''\cite{safety_tax}. Technical AI safety research progress could be slow. Connecting AI systems directly to the Internet and allocating greater computing resources will make them more capable and commercially attractive. In a competitive environment, this means that unsafe systems will be favored. Though we currently lack substantial empirical evidence supporting safer architectures, there appears to be a theoretical possibility of developing systems with clear boundaries and coordination to prevent alternative forms of development.\footnote{For instance, one approach for creating safer AI systems is through `Cognitive Emulations', a method that tries to build predictably limited systems by mimicking human-like processes. Another naively `safe' system is the Open Agency model, which separates world models from data and planning and acting in real-time. More generally, from first principles, there do exist human-level systems that do not unilaterally result in human extinction: humans. All of these proposals share the property that these systems cannot compete with black box systems with increased amounts of compute.}

Furthermore, if not managed, the pursuit of safer architectures may be dangerous. For example, one strategy MAGIC might adopt involves designing constrained systems that reduce their dependence on black-box models. This would be accomplished by inventing methods to attain the same level of capability using smaller black-boxes. If these techniques were to be released globally, they could potentially be applied to the largest models currently available, resulting in significantly enhanced capabilities. Given this, research undertaken at MAGIC should be thoroughly vetted before being shared with other researchers, including those from signatory countries.

\subsection{Secure: Among the Most Highly Secure Facilities on Earth}

\textbf{MAGIC would be one of the most secure facilities on Earth.} Security is crucial due to the immense potential for misuse, the risk of powerful models leaking by accident, and interest of malicious actors in acquiring the valuable technological artifacts that MAGIC would produce. Even a system designed with safety measures could pose a threat if exploited by malicious actors, and this security must be rigorously upheld through digital, physical, and personnel measures.

\textbf{With sufficient motivation and resources, secret information can stay secret.\footnote{The US military kept the nature of the Manhattan project secret from 1942-46, despite 130,000 people working on it at its peak.\cite{manhattan_project}}}Like nuclear weapons systems, MAGIC would have specialized hardware to remain isolated from the Internet. Further physical access devices could include biometric access controls, video surveillance, security patrols, and physical isolation of the facility. Scientists would operate on a compartmentalized basis, with strict background checks and lifestyle monitoring.\footnote{This is common in other projects. At the Wuhan Institute of Virology's BSL-4 lab, researchers are divided into groups that operate independently;  CERN scientists working with classified information require SECRET-level clearance.  The IAEA monitors scientists at nuclear facilities to identify insider threats early, including scrutinizing finances, psychological fitness, and personal connections.} Digital protections would ensure that sensitive systems are air-gapped from external networks and use specialized encrypted hardware.\footnote{ITER has an isolated control system secured through cryptography.}Scientists working at MAGIC would be required to sign strict non-disclosure agreements, protecting classified information under threat of legal consequences, including imprisonment. Additionally, non-compete clauses would ensure these scientists do not engage in similar AI projects after leaving MAGIC. It would additionally need to be located in a geopolitically and geographically neutral and secure location.\footnote{ One example of a location for such a facility would be in a country like Singapore because it has technological expertise, technocratic governance, geopolitical stability, and a strong relationship with both the US and China.}

\textbf{\textit{Limitations:}} Centralization and suppression of AI safety research could prove to be inefficient and limit innovation. A diversity of approaches and research groups would likely yield faster progress in advanced AI safety. Additionally, while hardware resources could be consolidated, it would be unnecessary to physically relocate researchers to a single site given the remoteness of the work. MAGIC can, however, take a more adaptable approach, involving a small (<10) network of coordinated labs, each with independent scientific direction, though this would present further tradeoffs for security and control. \cite{bengio_economist, bengio_blog}

\subsection{Collective: Supported Internationally, Benefits Distributed To All Signatories}

\textbf{MAGIC would be a multinational, cooperative effort enacted through an international treaty, with the goal of reducing risks from advanced AI systems.} The project could be championed by individual powerful states like the U.S. and China, along with an initial coalition of supporting countries. It would, however, also need to make early credible commitments that all other nations can participate. Inclusivity is important to ensure benefits of AI advancement such as scientific breakthroughs are shared equitably among nations, and ensure that a global moratorium is effective in practice.

\textbf{MAGIC's inclusive membership would be a core incentive for states. States which would otherwise be sidelined in a technology race can meaningfully influence the direction of AI development.} MAGIC would further give those states an opportunity to partake in the benefits of safe, powerful AI is developed. Currently, individual developing states and emerging market economies face a significant challenge in reining in the behaviors of powerful global technology corporations on their own. Meanwhile, they are likely to suffer some of the greatest damage of dangerous AI systems. Through inclusive membership, MAGIC is already an improvement compared to the current closed-door, highly informationally asymmetric situation in which a small group of companies continue with fast-paced, unfettered development.

\textbf{Inclusivity also ensures that MAGIC is viewed as a global effort for humanity's benefit by institutionalizing benefit-sharing into its governance structure.} This would involve the creation of a mechanism to distribute the benefits of AI research and development equitably among participating member states. We expect the first outputs of MAGIC to be scientific breakthroughs, such as an extension of the capabilities achieved by Alphafold. \cite{nature_alphafold} This is because as a secure global research institution, MAGIC can begin by focusing on narrow AI systems focused on solving specific scientific tasks, such as medical breakthroughs like a cure to cancer or longevity. The outputs of these would have to be verified to pose no safety risks before being releaseds. Because this information can be provided at low or zero marginal cost to an additional actor, MAGIC’s advancements can be distributed to all members equitably, acting as a further incentive. 

\textbf{After automation of scientific research in narrow domains, if MAGIC’s systems are deemed to be sufficiently safe, research could move onto general-purpose, powerful AI systems.} Once these controllable systems are achieved, it is likely they could be used to generate enormous economic value, as they will be able to automate a large fraction of economic tasks currently carried out by people. If these are deployed by member states, profits and economic gains from these applications will necessitate equitable economic distribution.

\textbf{\textit{Limitations:}} In practice, inclusion may pose a direct tradeoff with security. Though a globally inclusive recruitment model is ideal in theory, pragmatic concerns around preventing misuse of advanced AI may necessitate more selective participation criteria in practice. The goal should be assembling a world-class multinational team of AI experts, but with safeguards to keep sensitive research under control. However, this tradeoff can be mitigated through governance structures that enable broad participation in the moratorium, but narrow participation in the most advanced research.

Achieving this balance would require a rigorous selection process for candidates to join the research collective based on AI expertise, while also enabling broad international representation. This selection process would need to consider: a) necessary level of expertise b) conflicts of interest c) security checks d) international representation as possible. This would be operable because there are still strong incentives for nations to join even if they are not given access to MAGIC’s internal research, including economic benefits and greater global safety. The broader the participation, the more effective the moratorium would become.

\section{Conclusion}

MAGIC has precedence as an institution. When scientists discovered chlorofluorocarbons (CFCs) in man-made chemicals were creating a hole in the ozone layer, which posed a significant threat to human health and the environment, the international community agreed to phase out their production. The global community shifted from CFCs to the safer HFC technology through the Montreal Protocol which is considered the UN’s first universally ratified treaty.\footnote{ Additionally, in the case of unleaded gasoline, policy interventions led to the deliberate restriction and eventual phasing out of the incumbent technology – leaded gasoline. This facilitated the rise and development of a safer alternative, unleaded gasoline, which was initially less competitive but eventually surpassed the performance of leaded gasoline due to targeted advancements and societal preference for a less harmful option. The same can be seen in other industries, such as synthetic colourings in food production.} On matters of information security, the U.S. military maintained strict secrecy around the Manhattan Project from 1942 to 1946, despite involving approximately 130,000 people at its peak\cite{manhattan_project}. Additionally, the Trade-Related Aspects of Intellectual Property Rights (TRIPS) Agreement has allowed developing countries to access and produce lifesaving medicines researched and developed in more economically advanced countries. \cite{hoen_2002} These historical examples show that in the face of comparable global health or security risks, countries have come together to restrict the proliferation of dangerous technologies and carefully steward their safe and equitable development. 

MAGIC would not hinder the majority of current AI development. Narrow systems focused on specific domains like medical imaging or fraud detection would remain unaffected, and could continue to benefit economies and technological progress. While this proposal does not seek to minimize or dismiss the misuse risks of models below the frontier threshold, these models take supervised actions, not autonomous ones, and are largely controllable through human oversight and evaluation.

This paper does not delve into the political feasibility of implementing such an international regime or address the specific legislative strategies and rules that would need to be put in place to enforce a ban on high-capacity AGI training runs. We envision this process could plausibly occur through unilateral enforcement of national moratoriums in the United States and a small group of other nations with oversight of the majority of advanced AI development operations. Collectively, these national moratoria could serve as the foundation for an international treaty, and the basis for creating MAGIC. 

This paper does not seek to endorse MAGIC as the only solution, but rather one possible scenario which can effectively halt unchecked advanced AI advancement and significantly reduce large-scale risks of advanced AI. Governance models need not be mutually exclusive. It seems inevitable that advanced AI will require international governance at numerous levels, and other models including a scientific observatory\footnote{ Other proposals have called for an Intergovernmental Panel on Climate Change model for AI. 
} or an auditing mechanism\footnote{ Similar to the The International Atomic Energy Agency.
} could be complementary to MAGIC. \cite{ho_2023}

As AI capabilities accelerate, the risks of uncontrolled development also multiply. This paper outlines MAGIC as one potential governance framework to institute inclusive oversight and responsible development of advanced AI models. By consolidating the most advanced research within a single international body, MAGIC can enact a global moratorium on unauthorized models while enabling safety-focused progress within tight parameters. The aim is not absolute prohibition, but careful stewardship. With coordinated action, the potential of advanced AI models can be harnessed and equitably distributed, while risks are monitored and managed.

\printbibliography 
\end{document}